# SILM: A Subjective Intent Based Low-Latency Framework for Multiple Traffic Participants Joint Trajectory Prediction

Qu Weiming[1, †], Wang Jia[1, †], Du Jiawei[1], Zhu Yuanhao[3], Yu Jianfeng[3], Xia Rui[3], Cao Song[3], Wu Xihong[1, 2], Luo Dingsheng[1, 2, *]
[1]National Key Laboratory of General Artificial Intelligence, Key Laboratory of Machine Perception (MoE),
School of Intelligence Science and Technology, Peking University, Beijing 100871, China.
[2]PKU-WUHAN Institute for Artificial Intelligence.
[3]China Automotive Innovation Corporation
[†] These authors contributed equally to this work and should be considered co-first authors.
* Corresponding author
Email: dsluo@pku.edu.cn

*Abstract—* Trajectory prediction is a fundamental technology for advanced autonomous driving systems and represents one of the most challenging problems in the field of cognitive intelligence. Accurately predicting the future trajectories of each traffic participant is a prerequisite for building high safety and high reliability decision-making, planning, and control capabilities in autonomous driving. However, existing methods often focus solely on the motion of other traffic participants without considering the underlying intent behind that motion, which increases the uncertainty in trajectory prediction. Autonomous vehicles operate in real-time environments, meaning that trajectory prediction algorithms must be able to process data and generate predictions in real-time. While many existing methods achieve high accuracy, they often struggle to effectively handle heterogeneous traffic scenarios. In this paper, we propose a Subjective Intent-based Low-latency framework for Multiple traffic participants joint trajectory prediction. Our method explicitly incorporates the subjective intent of traffic participants based on their key points, and predicts the future trajectories jointly without map, which ensures promising performance while significantly reducing the prediction latency. Additionally, we introduce a novel dataset designed specifically for trajectory prediction. Related code and dataset will be available soon.

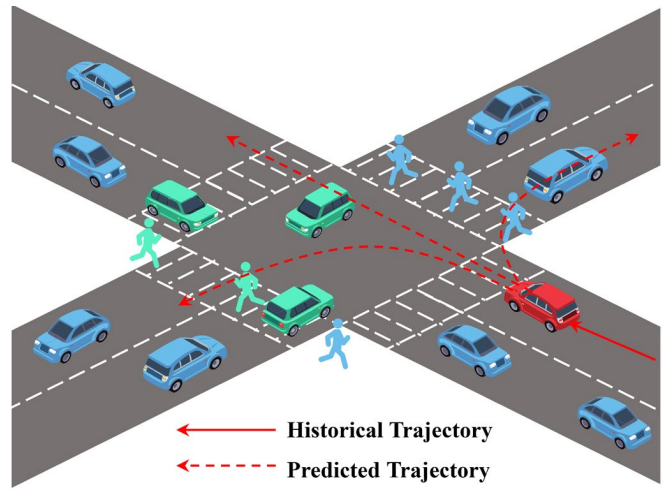

Fig.1. Depiction of the challenges of trajectory prediction. The Ego-Vehicle is represented by the red vehicle. **Multimodality**: In this scenario, despite having only one single historical trajectory, the Ego-Vehicle has three potential future legal trajectories. **Interdependence**: If the Ego-Vehicle chooses to make a left turn, to ensure socially acceptable driving behavior, it must consider the future movements of the green vehicle and the green pedestrian. **Real-time Requirement**: Due to the dynamic nature of driving scenarios, the Ego-Vehicle must respond in real-time.

## I. INTRODUCTION

The World Health Organization (WHO) published a report stating that in 2021, the number of traffic accident fatalities was 1.19 million, equivalent to 15 deaths per 100,000 people, with one person dying every 24 seconds due to traffic accidents [1]. These accidents also result in significant economic losses for individuals, families, and entire nations, amounting to 3% of the Gross Domestic Product (GDP) of most countries [2]. A study by National Highway Traffic Safety Administration (NHTSA) in the United States reveals that approximately 94% of severe traffic accidents are caused by driver errors [3]. Consequently, autonomous vehicles (AVs) are expected to play a crucial role in reducing traffic accidents and enhancing road safety.

Current autonomous driving systems typically divide the problem into four steps: perception, tracking, trajectory prediction, and path planning [4]. Trajectory prediction plays a pivotal role in autonomous driving. It serves not only as an extension of environmental perception and a basis for decision-making and planning, but also as the foundation for risk assessment and safety enhancement. Autonomous vehicles should be capable of predicting the future state of the surrounding environment in real-time, similar to human drivers. Based on this prediction, the vehicle can initiate new driving actions, such as acceleration or lane changes, to better adapt to the dynamic traffic environment.

In recent years, significant progress has been made in autonomous vehicle trajectory prediction, such as QCNet [5], BAT [6], LAformer [7]. However, accurately predicting the future trajectories of traffic participants still presents several challenges, as illustrated in Fig. 1. **Multimodality**: A past trajectory can have multiple potential future trajectories. Therefore, trajectory prediction models must learn to capture the underlying multimodal distribution, rather than simply predicting the most common pattern. Many studies leverage handcrafted anchor points to guide multimodal predictions [8] [9][10]. However, the effectiveness of these methods is highly dependent on the quality of the anchor points. Other works attempt to avoid this limitation by directly predicting multiple trajectories [11][12][13], which introduces the risk of mode collapse and training instability [5]. **Interdependence**: In heterogeneous road environments, such as intersections, campuses, or areas with mixed pedestrian and vehicle traffic, strong interactions exist among different traffic participants. Consequently, their trajectories are interdependent. Predicting trajectories in such complex environments requires models capable of handling the game-theoretic interactions between multiple traffic participants. Current trajectory prediction methods typically focus solely on the motion of other traffic participants without considering the underlying context or intent behind that motion, which further increases the uncertainty of trajectory prediction [3]. **Real-time Requirement**: Although a series of models have achieved impressive performance on trajectory prediction benchmarks [14][15][16], they often fail to efficiently handle the heterogeneous traffic scenarios. Jointly optimizing the trajectories of multiple traffic participants typically requires a large deep learning architecture. However, autonomous vehicles need to operate in real-time, typically at around 10 Hz, which places stringent constraints on the runtime of the prediction module [4].

The analysis above drives us to propose a **s**ubjective **i**ntent-based **l**ow-latency framework for **m**ultiple traffic participants joint trajectory prediction, termed as **SILM**. Inspired by HiVT [13], we first use a local encoder to extract spatiotemporal features from each traffic participant, followed by a global encoder to aggregate these features across all participants. Unlike HiVT, our method operates without reliance on map information and adopts an ego-vehicle-centered representation, as we focus more on how the future trajectories of surrounding traffic participants affect the ego-vehicle's behavior. At the same time, recognizing the importance of subjective intent in trajectory prediction, we explicitly extract key point features from each traffic participant to capture their intent. These key point features are then integrated with the trajectory features to enhance the performance.

In summary, the main contributions of this paper are summarized as follows:

- We propose a novel trajectory prediction framework without map that significantly improves prediction performance while maintaining low latency.
- We incorporate subjective intention cues from traffic participants, demonstrating their critical role in improving trajectory prediction accuracy.
- We contribute a new, specialized dataset for trajectory prediction tasks.

## II. RELATED WORKS

### A. Physics-based Methods

Physics-based methods rely on physical principles, taking into account physical factors such as current position, velocity, acceleration, and road constraints for trajectory prediction [17]. Common physics models include dynamic models [18][19][20] and kinematic models [21][22][23]. However, in practical trajectory prediction, noise interference is often present. The Kalman Filter (KF) method models the noise in the current vehicle state and its physical model using a Gaussian distribution. By combining the prediction and updating steps into a loop, the average value and covariance matrix of the vehicle state at each future time step can be obtained, yielding an average trajectory with associated uncertainty [24]. When no assumptions of linearity or Gaussian properties are made, Monte Carlo methods can approximate the state distribution. They perform random sampling of input variables and apply a physical model to generate potential future trajectories [25]. While these methods offer interpretability and computational efficiency, they are only suitable for simple prediction scenes and typically exhibit lower accuracy [26]. To achieve SOTA performance, physics-based methods may need to be combined with learning-based methods [27][28].

### B. Deep Learning-based Methods

To overcome the limitations of physics-based methods, deep learning-based methods have gained significant attention. Sequential Networks, Graph Neural Networks (GNN), and Generative Models are among the most widely used methods. Sequential Networks, such as Recurrent Neural Networks (RNN), Convolutional Neural Networks (CNN), and Attention Mechanisms (AM), are employed to extract features from historical trajectories. RNNs, such as Long Short-Term Memory Network (LSTM) [29] and Gated Recurrent Unit (GRU) [30], are designed to handle temporal information, while CNNs can effectively process spatial information including the interaction-related factors between traffic participants [31]. Some researchers combine RNN and CNN to integrate both temporal and spatial information into their models, such as Social-LSTM [29] and Traphic [32]. AM mimics the way humans think and allows the human to use limited attention resources to quickly filter out high-value information [26], which is also demonstrated superior performance in trajectory prediction [33][34]. However, when considering the complex interactions among traffic participants, Sequential Networks often fail to deliver satisfactory results. GNNs, such as Graph Convolutional Network (GCN) and Graph Attention Network (GAT) can effectively address this issue [35]. In this case, each individual in the environment can be viewed as a node in a graph, with edges representing the relationships between them. To address the multimodality issue in trajectory prediction, a plenty of researchers utilize Generative Models to generate multi-modal trajectories, such as Generative Adversarial Network (GAN) and Conditional Variational Auto Encoder (CVAE) [36][37][38][39].

### C. Reinforcement Learning-based Methods

Reinforcement Learning (RL) provides a new way to handle complex high-dimensional strategies. When applied to trajectory prediction, Markov Decision Process (MDP) is typically used to maximize cumulative rewards. RL helps estimate the reward function or directly identify the best trajectory prediction strategy. In this framework, traffic participants are assumed to act optimally based on a specific reward function. Inverse Reinforcement Learning (IRL), instead of imitating trajectories directly, seeks to understand the motivations behind them by inferring the reward function, which is then used to predict future trajectories. Due to the complexity of traffic behaviors, manually defining the reward function is challenging. Xu et al. [40] introduced Inverse Optimal Control (IOC), which uses Langevin sampling to model vehicle cost functions. Deep IRL combines IRL with deep neural networks (DNN) to learn reward functions from data. Guo et al. [41] proposed a multimodal trajectory prediction framework using IRL, which learns in an end-to-end manner. However, IRL is often difficult to train in scenarios with limited or no direct reward functions. Imitation Learning (IL) effectively addresses this issue, as it mimics expert behavior to quickly find strategies. Common methods include Behavior Cloning [42] and Generative Adversarial Imitation Learning (GAIL) [43].

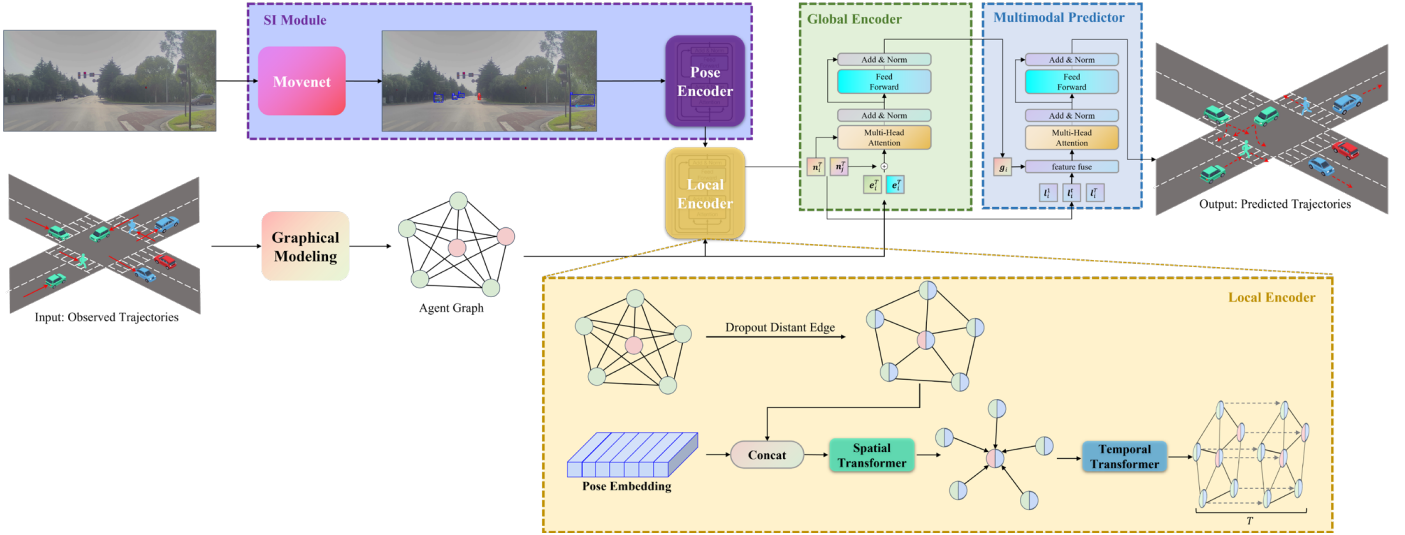

Fig.2. Workflow of SILM.

## III. PROBLEM FORMULATION

Accurate trajectory prediction of surrounding traffic participants necessitates a comprehensive understanding of the spatiotemporal dynamics of the environment, including the historical states of observable traffic participants and their interaction patterns. The historical states of traffic participants observed by the AVs is:

$$X = \{e_j^0, e_j^1, e_j^2, \ldots, e_j^{t-1}\}_{j=1}^N \tag{1}$$

where $N$ represents all traffic participants detected by the ego-vehicle; $t$ represents the length of the historical trajectory and $e^t$ denotes the state of the traffic participant at the current time. $X$ is the input of the prediction model, and traffic participants' trajectory with a time step length $f$ is predicted. This task can be mathematically expressed as the computation of the posterior distribution. If $I$ represents the environmental information, i.e., the road context, the output of the model can be expressed as:

$$Y = P(\{e_j^t, e_j^{t+1}, e_j^{t+2}, \ldots, e_j^{t+f}\}_{j=1}^N | X \cup I) \tag{2}$$

## IV. METHODOLOGY

We first introduce the general overview of SILM in Sec. IV.A. Next, we elaborate on the crucial design steps of the Subjective Intent (SI) Module in Sec. IV.B. After that, we delve into the details of the Local Encoder and the Global Encoder in Sec. IV.C and Sec. IV.D. Finally, we illustrate our Multimodal Predictor.

### A. Framework Overview

We propose a novel subjective intent based low-latency framework for joint trajectory prediction of multiple traffic participants, as illustrated in Fig.2. First, a local encoder is employed to extract spatiotemporal features for each traffic participant, capturing their motion patterns over time. At the same time, we explicitly extract key point features from each participant to capture their intent, which are then fused with trajectory features. Subsequently, a global encoder aggregates these local features across all participants, utilizing multi-head attention to model interactions and dependencies between different traffic participants. Finally, the multimodal predictor combines these features to generate multiple plausible future trajectories, offering a robust and flexible solution for real-time trajectory prediction in dynamic and heterogeneous traffic environments.

### B. Subjective Intent Module

To explicitly infer motion intent, our SI module utilizes a Movenet-based detector to extract category-specific key points from sequential observations of traffic participants.

For pedestrians, we detect nine biomechanical key points (face, shoulders, hips, knees, ankles) to infer body orientation and walking intent. For vehicles, we identify nine geometric key points (center, corners, wheels) that capture heading and acceleration cues, while bicycles and motorcycles are represented by nine dynamic points (handlebars, seat, pedals/footrests, wheels) to model rider pose and turning intent.

These key points are processed through a Transformer encoder module to generate pose feature embeddings ($Z_{pose}$), which are then concatenated with the trajectory embeddings ($Z_{tra}$). This fusion enables the intent features to assist in trajectory reasoning, thereby enhancing the accuracy of trajectory prediction in a computationally efficient manner.

### C. Local Encoder

In the graph constructed at each timestamp $t$, the original feature representation of node $i$ is as follows:

$$n_i^t = \{\Delta x_i^t, \Delta y_i^t, \dot{x}_i^t, \dot{y}_i^t, \theta_i^t\} \tag{3}$$

where $\Delta x_i^t = x_i^t - x_i^{t-1}$ and $\Delta y_i^t = y_i^t - y_i^{t-1}$, $\dot{x}_i^t$ and $\dot{y}_i^t$ denote velocity, and $\theta_i^t$ is the yaw angle.

We then apply a simple Multi-Layer Perceptron (MLP) to map the trajectory features to the same dimension $d_h$. After that, the trajectory embeddings and pose feature embeddings are concatenated.

$$z_{tra_i}^t = \phi_{mlp}(n_i^t) \tag{4}$$

$$z_{con_i}^t = \phi_{concat}(\{z_{tra_i}^t, z_{pose_i}^t\}) \tag{5}$$

To enhance computational efficiency, during the local feature update phase, only the interaction relationships between the central agent and those within a specified surrounding range are considered (the radius is set to $\gamma$) and the connection edges to the central agent are removed for targets that are too distant. To effectively model the interactions within a certain range around agent $i$, we employ multiple attention modules. In these modules, $z_{con_i}^t$ serves as the query vector, while the feature vectors $z_{con_{S_i}}^t \in \mathbb{R}^{d_a \times 2d_h}$ ($S_i$ is the set of surrounding agents, $d_a$ donates the number of the surrounding agents) act as the key and value vectors:

$$q_i^t = z_{con_i}^t W_{spa}^Q, \ k_{S_i}^t = z_{con_{S_i}}^t W_{spa}^K, \ v_{S_i}^t = z_{con_{S_i}}^t W_{spa}^V \quad (6)$$

$$AW_{spa_i}^t = softmax(\frac{q_i^t}{\sqrt{d_h}} \cdot k_{S_i}^{t\ T}) \cdot v_{S_i}^t \quad (7)$$

$$z_{spa_i}^t = \alpha \cdot z_{con_i}^t + (1-\alpha) \cdot AW_{spa_i}^t \quad (8)$$

$$Z_{spa_i} = \{z_{spa_i}^t\}_{t=1}^T \quad (9)$$

where $W_{spa}^Q$, $W_{spa}^K$, $W_{spa}^V \in \mathbb{R}^{2d_h \times 2d_h}$ are learnable matrices, and $\alpha$ is predefined hyperparameters.

After aggregating the local spatial information of each agent, we further apply a Transformer Encoder module to learn the temporal evolution of the agents' motion patterns:

$$Q_i = Z_{spa_i} W_{tem}^Q, \ K_i = Z_{spa_i} W_{tem}^K, \ V_i = Z_{spa_i} W_{tem}^V \quad (10)$$

To better handle sequential features, a mask is employed to ensure that the model focuses solely on the known portion at each step during processing.

$$Z = softmax(\begin{cases} \frac{Q_i}{\sqrt{d_h}} \cdot K_i, & u \le v \\ -inf, & u > v \end{cases}) \cdot V_i \quad (11)$$

### D. Global Encoder

To capture global scene information, the global interaction module is designed to learn the relative interaction relationships among all agents. Similar to HiVT [13], the primitive feature of the edge connecting two adjacent nodes is defined as follows:

$$e_{ij}^{T_L} = \{\Delta x_{ij}^{T_L}, \Delta y_{ij}^{T_L}, \dot{x}_{ij}^{T_L}, \dot{x}_{ij}^{T_L}, \cos(\Delta\theta)_{ij}^{T_L}, \sin(\Delta\theta)_{ij}^{T_L}\} \quad (12)$$

which represents the relative pose between the two agents at the last observed time step $T_L$. $e_{ij}^{T_L}$ is then mapped to a $d_h$-dimensional space through MLPs.

$$q_i = W_{glo}^Q z_{loc_{ij}}^{T_L}, \ k_{ij} = W_{glo}^K[z_{loc_{ij}}^{T_L}, e_{ij}^{T_L}], \ v_{ij} = W_{glo}^V[z_{loc_{ij}}^{T_L}, e_{ij}^{T_L}] \quad (13)$$

$$z_{glo_i}^{T_L} = softmax(\frac{q_i}{\sqrt{d_h}} \cdot k_{ij}) \cdot v_{ij} \quad (14)$$

### E. Multimodal Predictor

Since our method does not incorporate map structural information, additional uncertainty is introduced during the trajectory decoding phase to compensate for this information deficit and enhance the model's generalization capability. The input to the predictor consists of local features and global features. Initially, the global features are passed through an MLP layer to generate a fusion parameter tensor $\{\delta_i^t\}_{i=1 \sim N}^{T_L}$. The local and global features are then fused as follows:

$$z_i^t = \delta_i^t \cdot z_{loc_i}^t + (1 - \delta_i^t) \cdot z_{glo_i}^{T_L} \quad (15)$$

Subsequently, a Transformer Decoder module is employed to iteratively generate the predicted trajectory points. The output is a set of trajectory points with dimensions $N \times M \times T_f \times 4$, where $N$ represents the number of the agents, $M$ denotes the number of output modes, and $T_f$ represents the number of prediction time steps. The last dimension of the output feature corresponds to the relative $x$ and $y$ coordinates with respect to the ego-vehicle, as well as scaling parameters in the $x$ and $y$ dimensions.

Like most trajectory prediction methods, the optimization objective of our method is composed of two parts: classification loss ($loss_{cls}$) and regression loss ($loss_{reg}$):

$$loss_{cls} = -\frac{1}{N}\sum_{i=1}^{N}\sum_{j=1}^{M} y_{i,j} \log(p_{i,j}) \quad (16)$$

$$loss_{reg} = \frac{1}{N}\sum_{i=1}^{N}\sum_{t=1}^{T_f} \|(p_i^t - \hat{p}_i^t)\|_2 \quad (17)$$

$$loss = loss_{cls} + loss_{reg} \quad (18)$$

## V. EXPERIMENTS

### A. Environmental Setup

**Datasets.** We collect a new dataset for trajectory prediction tasks, termed as CAIC-TP. The dataset comprises more than 25000 sequences of data collected from Nanjing, China, sampled at a rate of 2Hz. In addition, the proposed method is also developed and evaluated on other two widely used datasets for autonomous driving: Argoversev2 End-to-End (E2E) Forecasting [44] and NuScenes [45]. To ensure fair comparisons, our method and baseline methods are evaluated under identical experimental setting. Models are required to predict agents' 3-second future trajectories given the 2-second historical observations. By leveraging these three datasets, we aim to thoroughly evaluate the models' prediction capability on various data distributions.

**Metrics.** We adopt the standard evaluation metrics to measure the trajectory prediction performance, including minimum Average Displacement Error (minADE), minimum Final Displacement Error (minFDE), weighted sum of Average Displacement Error (WSADE), weighted sum of Final Displacement Error (WSFDE), Miss Rate (MR), and Computation Time (Speed). Relevant formulas are defined as follows:

$$ADE = \frac{1}{N}\sum_{i=1}^{N}(\frac{1}{T}\sum_{t=1}^{T}\|P_{i,t}^k - G_{i,t}\|^2) \quad (19)$$

$$FDE = \frac{1}{N}\sum_{i=1}^{N}\|P_{i,T}^k - G_{i,T}\|^2 \quad (20)$$

$$minADE = \frac{1}{N}\sum_{i=1}^{N} min_k(\frac{1}{T}\sum_{t=1}^{T}\|P_{i,t}^k - G_{i,t}\|^2) \quad (21)$$

$$minFDE = \frac{1}{N}\sum_{i=1}^{N} min_k \|P_{i,T}^k - G_{i,T}\|^2 \quad (22)$$

$$WSADE = D_v \cdot ADE_v + D_p \cdot ADE_p + D_b \cdot ADE_b \quad (23)$$

$$WSFDE = D_v \cdot FDE_v + D_p \cdot FDE_p + D_b \cdot FDE_b \quad (24)$$

$$MR = \frac{1}{N}\sum_{i=1}^{N} \mathbb{1}(min_k \|P_{i,T}^k - G_{i,T}\|^2 > \tau) \quad (25)$$

where $N$ represents the total number of samples, $T$ denotes the prediction time steps, $P_{i,t}^k$ is the $k^{th}$ predicted position of the $i^{th}$ sample at time $t$, and $G_{i,t}$ is the ground truth position of the $i^{th}$ sample at time $t$. The weighted parameter $D_v$, $D_p$, $D_b$ are inversely proportional to the average speeds of vehicles, pedestrians, and bicyclists respectively. The indicator function $\mathbb{1}(\cdot)$ outputs 1 if the condition in parentheses is satisfied, and 0 otherwise. $\tau$ represents the prediction error threshold.

**Implementation Details.** In our experiments, we used two RTX 3090 GPUs for model training, with a batch size of 32. For inference speed testing, a single GPU device was employed. The Adam optimizer was used to optimize training performance, with a learning rate of $5 \times 10^{-4}$ and an initial weight decay rate of $1 \times 10^{-4}$. To ensure a fair comparison, the embedding dimension was consistently set to 64 across all methods, and all models were trained for 128 epochs. The hyperparameters for SLIM are presented in Table I.

TABLE I. HYPERPARAMETERS SETTINGS

| Hyperpara-meters | Learning Rate | Weight decay | Batch Size | Epoch | $d_h$ | $\alpha$ | $\gamma$ |
|---|---|---|---|---|---|---|---|
| Value | 0.0005 | 0.0001 | 32 | 128 | 64 | 0.5 | 40 |

*B. Quantitative Results*

Based on the experimental results in Table II, the proposed SILM framework demonstrates significant advantages across multiple traffic participant trajectory prediction tasks. While LT3D achieves the fastest computation speed, it exhibits notably poor performance across other metrics. In contrast, SILM outperforms baseline methods on the Argoversev2 dataset, excelling in both minADE and minFDE, while maintaining a competitive processing speed. On our CAIC-TP dataset, SILM further demonstrates its robustness, showing improvements in minADE, WSADE, and WSFDE, underscoring its strong generalization ability in complex, heterogeneous traffic scenarios. On the NuScenes dataset, SILM sets new records for the evaluation metrics, establishing it as the best-performing model in this dynamic and high-traffic environment. These results confirm that SILM successfully balances multimodal prediction capabilities with real-time requirements, providing an efficient and reliable solution for trajectory prediction in multi-participant traffic scenarios.

*C. Qualitative Results*

We present a qualitative analysis of SILM over three datasets to demonstrate its effectiveness in trajectory prediction. As shown in Fig. 3, our model successfully handles complex traffic scenarios, such as intersections, pedestrian-vehicle interactions, and mixed traffic conditions, where traditional models often struggle. To highlight the impact of the SI Module, we visualize the key points of the traffic participants. By incorporating the subjective intent of traffic participants, our model is able to generate multimodal and realistic future trajectories, effectively capturing the diverse range of potential behaviors of each participant. This capability of predicting multiple plausible future trajectories allows our model to account for the inherent uncertainty in human driving behavior. Furthermore, the simultaneous prediction of trajectories for multiple traffic participants enables our model to better simulate real-world interactions and anticipate potential conflicts or cooperation among agents.

TABLE II. PERFORMANCE COMPARISON OF DIFFERENT METHODS OVER THREE DATASETS

| Dataset | Method | Year | Venue | minADE$_6$ | minFDE$_6$ | WSADE | WSFDE | MR | Speed(ms) |
|---|---|---|---|---|---|---|---|---|---|
| Argoversev2 | LT3D | 2022 | CoRL | 3.6338 | 4.0969 | 2.8888 | 2.8681 | 0.3870 | **0.0901** |
| | HiVT | 2022 | CVPR | 0.1021 | 0.2492 | **0.0769** | 0.2101 | 0.0128 | 0.4576 |
| | Wayformer | 2023 | ICRA | 1.0119 | 1.7390 | 0.8432 | 1.6242 | 0.3267 | 3.4389 |
| | QCNet | 2023 | CVPR | 0.0979 | 0.2535 | 0.0789 | **0.2094** | 0.0119 | 0.4338 |
| | LAFormer | 2024 | CVPR | 0.3509 | 0.5238 | 0.2599 | 0.4016 | 0.0234 | 5.6670 |
| | **Ours** | 2025 | IROS | **0.0916** | **0.2378** | 0.0799 | 0.2171 | **0.0096** | 0.3983 |
| CAIC-TP | LT3D | 2022 | CoRL | 6.1607 | 9.1894 | 2.7466 | 3.4299 | 0.4746 | **0.0379** |
| | HiVT | 2022 | CVPR | 0.8039 | **1.3132** | 0.4795 | 0.7267 | **0.0824** | 0.1149 |
| | Wayformer | 2023 | ICRA | 1.5843 | 2.9577 | 0.5423 | 0.9952 | 0.2077 | 1.1394 |
| | QCNet | 2023 | CVPR | 1.3336 | 2.2188 | 0.9302 | 1.5550 | 0.1186 | 0.2245 |
| | LAFormer | 2024 | CVPR | 0.8651 | 1.3976 | 0.6459 | 0.9548 | 0.2344 | 0.9228 |
| | **Ours** | 2025 | IROS | **0.7760** | 1.4051 | **0.4102** | **0.6707** | 0.1515 | 0.2393 |
| NuScenes | LT3D | 2022 | CoRL | 2.6756 | 3.9022 | 1.9364 | 2.6191 | 0.3049 | **0.0374** |
| | HiVT | 2022 | CVPR | 0.2044 | 0.3432 | 0.1677 | 0.2751 | 0.0305 | 0.0930 |
| | Wayformer | 2023 | ICRA | 0.8490 | 1.5517 | 0.5264 | 0.9438 | 0.2171 | 1.4290 |
| | QCNet | 2023 | CVPR | 0.2136 | 0.3791 | 0.1799 | 0.3144 | 0.0387 | 0.1869 |
| | LAFormer | 2024 | CVPR | 0.6815 | 1.2779 | 0.3840 | 0.7208 | 0.1484 | 0.8153 |
| | **Ours** | 2025 | IROS | **0.1962** | **0.3300** | **0.1082** | **0.1692** | **0.0287** | 0.1135 |

Best results are **boldened**, and worst results are underlined.

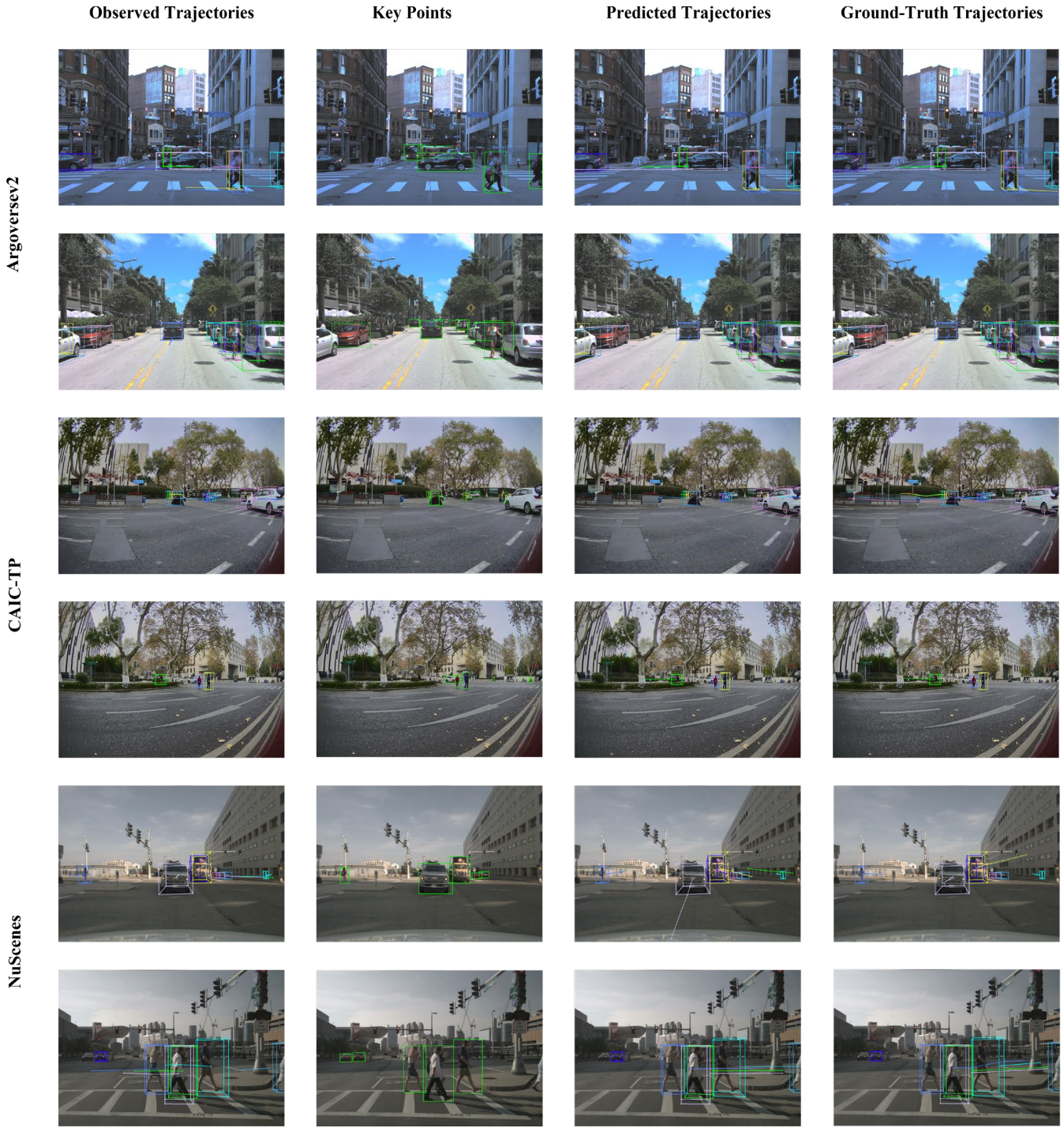

Fig.3. Qualitative results of SILM over three datasets. For clarity, we only visualize the trajectories with the highest confidence.

## D. Ablation Studies

We also conduct an ablation analysis over the three datasets to assess the relative importance of the SI Module in our trajectory prediction framework. The results of each model setting are summarized in Table III. Our analysis reveals that incorporating the SI Module leads to a consistent and significant improvement in model performance across all evaluated metrics, including minADE, minFDE, WSADE and WSFDE. Specifically, by explicitly considering the subjective intent of traffic participants, the model is able to better capture the underlying motivations behind their movements, leading to more accurate and reasonable trajectory predictions. This enhancement is particularly noticeable in scenarios involving complex and dynamic traffic environments, where understanding the intent behind motion is crucial for reducing uncertainty in future trajectory prediction. In summary, the addition of the SI Module not only boosts the accuracy of trajectory prediction but also contributes to a more robust model.

TABLE III. Ablation Studies of the SI Module in SILM

| Dataset | SI | minADE$_6$/minFDE$_6$ | WSADE/WSFDE |
|---|---|---|---|
| Argoversev2 | ✘ | 0.1350/0.3087 | 0.0917/0.2407 |
| | ✔ | **0.0916/0.2378** | **0.0799/0.2171** |
| CAIC-TP | ✘ | 1.2336/2.0077 | 0.6112/0.8286 |
| | ✔ | **0.7760/1.4051** | **0.4102/0.6707** |
| Nu-Scenes | ✘ | 0.2175/0.3741 | 0.1902/0.3242 |
| | ✔ | **0.1962/0.3300** | **0.1082/0.1692** |

## VI. Conclusion

In this paper, we propose SILM, a Subjective Intent-based Low-latency framework for joint trajectory prediction, which addresses the critical challenges of balancing accuracy, real-time performance, and generalization in autonomous driving. By explicitly modeling traffic participants' intent through key point features and integrating these cues with spatiotemporal dynamics, SILM eliminates reliance on HD maps while achieving satisfying performance across benchmarks. These results validate SILM's ability to harmonize intent-aware prediction with real-time efficiency in heterogeneous scenarios. SILM's map-free, ego-centric design ensures scalability in unstructured environments, while its interpretable intent modeling reduces prediction uncertainty.

## Acknowledgment

The work is supported in part by the National Natural Science Foundation of China (No. 62176004, No. U1713217), Intelligent Robotics and Autonomous Vehicle Lab (RAV), Wuhan East Lake High-Tech Development Zone National Comprehensive Experimental Base of Governance of Intelligent Society, and High-performance Computing Platform of Peking University.

## References

[1] World Health Organization. Global status report on road safety 2023. [EB/OL].https://www.who.int/publications/i/item/9789240086517, [2024-10-03].

[2] World Health Organization. Global Plan for the Decade of Action for Road Safety 2021-2030. [EB/OL]. https://www.who.int/publications/m/item/global- plan-for-the-decade-of-action-for-road-safety-2021-2030, [2024-10-03].

[3] Bharilya V, Kumar N. Machine learning for autonomous vehicle's trajectory prediction: A comprehensive survey, challenges, and future research directions[J]. Vehicular Communications, 2024: 100733.

[4] Singh A. Trajectory-Prediction with Vision: A Survey[C]//Proceedings of the IEEE/CVF International Conference on Computer Vision. 2023: 3318-3323.

[5] Zhou Z, Wang J, Li Y H, et al. Query-centric trajectory prediction[C]//Proceedings of the IEEE/CVF Conference on Computer Vision and Pattern Recognition. 2023: 17863-17873.

[6] Liao H, Li Z, Shen H, et al. Bat: Behavior-aware human-like trajectory prediction for autonomous driving[C]//Proceedings of the AAAI Conference on Artificial Intelligence. 2024, 38(9): 10332-10340.

[7] Liu M, Cheng H, Chen L, et al. Laformer: Trajectory prediction for autonomous driving with lane-aware scene constraints[C]//Proceedings of the IEEE/CVF Conference on Computer Vision and Pattern Recognition. 2024: 2039-2049.

[8] Chai Y, Sapp B, Bansal M, et al. MultiPath: Multiple Probabilistic Anchor Trajectory Hypotheses for Behavior Prediction[C]//Conference on Robot Learning. PMLR, 2020: 86-99.

[9] Phan-Minh T, Grigore E C, Boulton F A, et al. Covernet: Multimodal behavior prediction using trajectory sets[C]//Proceedings of the IEEE/CVF conference on computer vision and pattern recognition. 2020: 14074-14083.

[10] Zeng W, Liang M, Liao R, et al. Lanercnn: Distributed representations for graph-centric motion forecasting[C]//2021 IEEE/RSJ International Conference on Intelligent Robots and Systems (IROS). IEEE, 2021: 532-539.

[11] Cui H, Radosavljevic V, Chou F C, et al. Multimodal trajectory predictions for autonomous driving using deep convolutional networks[C]//2019 international conference on robotics and automation (ICRA). IEEE, 2019: 2090-2096.

[12] Varadarajan B, Hefny A, Srivastava A, et al. Multipath++: Efficient information fusion and trajectory aggregation for behavior prediction[C]//2022 International Conference on Robotics and Automation (ICRA). IEEE, 2022: 7814-7821.

[13] Zhou Z, Ye L, Wang J, et al. Hivt: Hierarchical vector transformer for multi-agent motion prediction[C]//Proceedings of the IEEE/CVF Conference on Computer Vision and Pattern Recognition. 2022: 8823-8833.

[14] Chang M F, Lambert J, Sangkloy P, et al. Argoverse: 3d tracking and forecasting with rich maps[C]//Proceedings of the IEEE/CVF conference on computer vision and pattern recognition. 2019: 8748-8757.

[15] Ettinger S, Cheng S, Caine B, et al. Large scale interactive motion forecasting for autonomous driving: The waymo open motion dataset[C]//Proceedings of the IEEE/CVF International Conference on Computer Vision. 2021: 9710-9719.

[16] Peri N, Dave A, Ramanan D, et al. Towards long-tailed 3d detection[C]//Conference on Robot Learning. PMLR, 2023: 1904-1915.

[17] Wong C, Xia B, Hong Z, et al. View vertically: A hierarchical network for trajectory prediction via fourier spectrums[C]//European Conference on Computer Vision. Cham: Springer Nature Switzerland, 2022: 682-700.

[18] Lin C F, Ulsoy A G, LeBlanc D J. Vehicle dynamics and external disturbance estimation for vehicle path prediction[J]. IEEE Transactions on Control Systems Technology, 2000, 8(3): 508-518.

[19] Kaempchen N, Schiele B, Dietmayer K. Situation assessment of an autonomous emergency brake for arbitrary vehicle-to-vehicle collision scenarios[J]. IEEE Transactions on Intelligent Transportation Systems, 2009, 10(4): 678-687.

[20] Brännström M, Coelingh E, Sjöberg J. Model-based threat assessment for avoiding arbitrary vehicle collisions[J]. IEEE Transactions on Intelligent Transportation Systems, 2010, 11(3): 658-669.

[21] Miller R, Huang Q. An adaptive peer-to-peer collision warning system[C]//Vehicular Technology Conference. IEEE 55th Vehicular Technology Conference. VTC Spring 2002 (Cat. No. 02CH37367). IEEE, 2002, 1: 317-321.

[22] Hillenbrand J, Spieker A M, Kroschel K. A multilevel collision mitigation approach—Its situation assessment, decision making, and performance tradeoffs[J]. IEEE Transactions on intelligent transportation systems, 2006, 7(4): 528-540.

[23] Lytrivis P, Thomaidis G, Amditis A. Cooperative path prediction in vehicular environments[C]//2008 11th International IEEE Conference on Intelligent Transportation Systems. IEEE, 2008: 803-808.

[24] Lefkopoulos V, Menner M, Domahidi A, et al. Interaction-aware motion prediction for autonomous driving: A multiple model kalman filtering scheme[J]. IEEE Robotics and Automation Letters, 2020, 6(1): 80-87.

[25] Wang Y, Liu Z, Zuo Z, et al. Trajectory planning and safety assessment of autonomous vehicles based on motion prediction and model predictive control[J]. IEEE Transactions on Vehicular Technology, 2019, 68(9): 8546-8556.

[26] Huang Y, Du J, Yang Z, et al. A survey on trajectory-prediction methods for autonomous driving[J]. IEEE Transactions on Intelligent Vehicles, 2022, 7(3): 652-674.

[27] Xie G, Gao H, Qian L, et al. Vehicle trajectory prediction by integrating physics-and maneuver-based approaches using interactive multiple models[J]. IEEE Transactions on Industrial Electronics, 2017, 65(7): 5999-6008.

[28] Song H, Luan D, Ding W, et al. Learning to predict vehicle trajectories with model-based planning[C]//Conference on Robot Learning. PMLR, 2022: 1035-1045.

[29] Alahi A, Goel K, Ramanathan V, et al. Social lstm: Human trajectory prediction in crowded spaces[C]//Proceedings of the IEEE conference on computer vision and pattern recognition. 2016: 961-971.

[30] Ding W, Chen J, Shen S. Predicting vehicle behaviors over an extended horizon using behavior interaction network[C]//2019 international conference on robotics and automation (ICRA). IEEE, 2019: 8634-8640.

[31] Phan-Minh T, Grigore E C, Boulton F A, et al. Covernet: Multimodal behavior prediction using trajectory sets[C]//Proceedings of the IEEE/CVF conference on computer vision and pattern recognition. 2020: 14074-14083.

[32] Chandra R, Bhattacharya U, Bera A, et al. Traphic: Trajectory prediction in dense and heterogeneous traffic using weighted interactions[C]//Proceedings of the IEEE/CVF Conference on Computer Vision and Pattern Recognition. 2019: 8483-8492.

[33] Liu Y, Zhang J, Fang L, et al. Multimodal motion prediction with stacked transformers[C]//Proceedings of the IEEE/CVF conference on computer vision and pattern recognition. 2021: 7577-7586.

[34] Zeng W, Li M, Xiong W, et al. Mpcvit: Searching for accurate and efficient mpc-friendly vision transformer with heterogeneous attention[C]//Proceedings of the IEEE/CVF International Conference on Computer Vision. 2023: 5052-5063.

[35] Diehl F, Brunner T, Le M T, et al. Graph neural networks for modelling traffic participant interaction[C]//2019 IEEE Intelligent Vehicles Symposium (IV). IEEE, 2019: 695-701.

[36] Gupta A, Johnson J, Fei-Fei L, et al. Social gan: Socially acceptable trajectories with generative adversarial networks[C]//Proceedings of the IEEE conference on computer vision and pattern recognition. 2018: 2255-2264.

[37] Li J, Ma H, Tomizuka M. Conditional generative neural system for probabilistic trajectory prediction[C]//2019 IEEE/RSJ International Conference on Intelligent Robots and Systems (IROS). IEEE, 2019: 6150-6156.

[38] Guo H, Meng Q, Zhao X, et al. Map-enhanced generative adversarial trajectory prediction method for automated vehicles[J]. Information Sciences, 2023, 622: 1033-1049.

[39] Yue J, Manocha D, Wang H. Human trajectory prediction via neural social physics[C]//European Conference on Computer Vision. Cham: Springer Nature Switzerland, 2022: 376-394.

[40] Xu Y, Zhao T, Baker C, et al. Learning trajectory prediction with continuous inverse optimal control via Langevin sampling of energy-based models[J]. arXiv preprint arXiv:1904.05453, 2019.

[41] Guo K, Liu W, Pan J. End-to-end trajectory distribution prediction based on occupancy grid maps[C]//Proceedings of the IEEE/CVF Conference on Computer Vision and Pattern Recognition. 2022: 2242-2251.

[42] Hu A, Corrado G, Griffiths N, et al. Model-based imitation learning for urban driving[J]. Advances in Neural Information Processing Systems, 2022, 35: 20703-20716.

[43] Bronstein E, Palatucci M, Notz D, et al. Hierarchical model-based imitation learning for planning in autonomous driving[C]//2022 IEEE/RSJ International Conference on Intelligent Robots and Systems (IROS). IEEE, 2022: 8652-8659.

[44] Peri N, Luiten J, Li M, et al. Forecasting from lidar via future object detection[C]//Proceedings of the IEEE/CVF Conference on Computer Vision and Pattern Recognition. 2022: 17202-17211.

[45] Caesar H, Bankiti V, Lang A H, et al. nuscenes: A multimodal dataset for autonomous driving[C]//Proceedings of the IEEE/CVF conference on computer vision and pattern recognition. 2020: 11621-11631.